\pdfoutput=1

\documentclass[11pt]{article}

\usepackage[final]{acl}

\usepackage{times}
\usepackage{latexsym}
\usepackage{amsmath} 
\usepackage[T1]{fontenc}

\usepackage[utf8]{inputenc}

\usepackage{microtype}

\usepackage{inconsolata}

\usepackage{graphicx}

\usepackage{xspace}
\newcommand{\proj}{UORA\xspace}
\newcommand{\project}{Uniform Orthogonal Reinitialization Adaptation\xspace}
\usepackage{multirow}
\usepackage{makecell}
\usepackage{xcolor}
\usepackage{booktabs}
\usepackage{amsfonts}
\usepackage{colortbl}
\usepackage{enumitem}
\usepackage{tcolorbox}
\usepackage{tabularx}
\usepackage[symbol]{footmisc}

\title{\proj: \project  in Parameter Efficient Fine-Tuning of Large Models}

  \author{Xueyan Zhang$^{1,*}$, Jinman Zhao$^{2,\dagger,*}$,  Zhifei Yang$^{3}$, Yibo Zhong$^{4}$,\\
  \bf Shuhao Guan$^{5}$, Linbo Cao$^{1},$Yining Wang$^{2}$\\
  $^1$University of Waterloo, $^2$University of Toronto,\\
  $^3$Peking University, $^4$Independent Researcher, $^5$University College Dublin\\
}

\begin{document}
\maketitle
\begin{abstract}
This paper introduces \project (\proj), a novel parameter-efficient fine-tuning (PEFT) approach for Large Language Models (LLMs). 
\proj achieves state-of-the-art performance and parameter efficiency by leveraging a low-rank approximation method to reduce the number of trainable parameters. 
Unlike existing methods such as LoRA and VeRA, \proj employs an interpolation-based reparametrization mechanism that selectively reinitializes rows and columns in frozen projection matrices, guided by the vector magnitude heuristic. 
This results in substantially fewer trainable parameters compared to LoRA and outperforms VeRA in computation and storage efficiency. 
Comprehensive experiments across various benchmarks demonstrate \proj's superiority in achieving competitive fine-tuning performance with negligible computational overhead. 
We demonstrate its performance on GLUE and E2E benchmarks and its effectiveness in instruction-tuning large language models and image classification models.
Our contributions establish a new paradigm for scalable and resource-efficient fine-tuning of LLMs.
\footnote[0]{$^{*}$Equal contribution}
\footnote[0]{$^{\dagger}$Corresponding author: jzhao@cs.toronto.edu}
\end{abstract}
\section{Introduction}
\begin{figure*}[!ht]
\centering
\includegraphics[width=\textwidth]{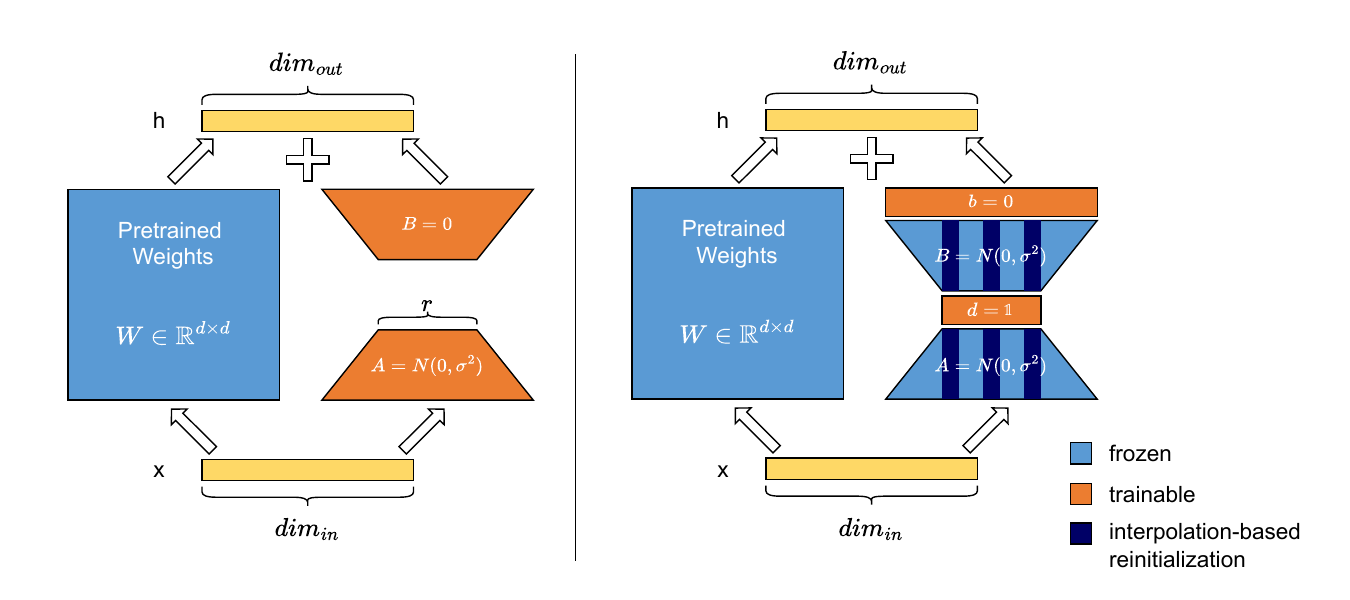}
\caption{\textbf{Overview of LoRA (left) and \proj (right).} LoRA trains a pair of projection matrices, namely A and B, with low rank $r$. The update to the pretrained weights is thus represented as ${\Delta} W = A \times B$. \proj adopts the similar strategy as VeRA; both projection matrices are frozen and randomized. A pair of scaling vectors, $\vec{d}$ and $\vec{b}$, is trained to adapt the frozen matrices. The key difference is that \proj applies interpolation reinitialization mechanism to selectively and partially update A and B. Similar to all LoRA-based PEFT methods, the learned weight update ${\Delta} W$ could be merged into $W$ for zero inference latency. 
}
\label{fig.overview}
\end{figure*}

Large models have demonstrated unparalleled performance in many tasks~\cite{li2024cpseg,zhang2025surveytesttimescalinglarge,wu2025cot,ma2025problemsolvinglogicguidedcurriculum}.
Popular models like GPT-3~\cite{brown2020language}, LLaMA3~\cite{dubey2024llama}, and Qwen~\cite{yang2024qwen2} have shown promising capability in various practical domains such as Reasoning~\cite{cai2025t2thoughtstemperaturetree,yang2024chainofthoughtlargelanguagemodels,xu2025redstardoesscalinglongcot,li202512surveyreasoning,bi2025cotkineticstheoreticalmodelingassessing}, education~\cite{chen-etal-2025-classic4children} and RAG~\cite{ref:septa,liu2024raginstructboostingllmsdiverse,ref:graphrag}.
However, many downstream applications still require fine-tuning to improve the zero-shot experience,
incorporate domain-specific content~\cite{wu2025iterpreffocalpreferencelearning,zheng2025asymmetric,luo2025beyond,wang2025texttocadgenerationinfusingvisual,dong2024generalizable}, and optimize performance in diverse scenarios~\cite{Chen_2024}. 
The full fine-tuning (FFT) process adapts the model by updating all pre-trained weights within the model. Depending on the goal of downstream tasks, FFT may require a significant amount of time and computing resources;
meanwhile, as the model size increases exponentially, the demand for multiple checkpoints for iterative pre-training and downstream adaption cycles poses significant challenges to storage efficiency. 

Parameter-efficient fine-tuning (PEFT) methods aim to address this challenge by injecting adapters
for the fine-tuning process while keeping the pre-trained weights untouched.
Compared to the existing approaches at the time~\cite{houlsby2019parameter, rebuffi2017learning, li2021prefix, lester2021power, hambardzumyan2021warp, liu2024gpt}, 
Low-Rank Adaptation (LoRA) exploits the low intrinsic dimensionality in weight matrices and performs low-rank approximation with a pair of projection-down and projection-up matrices; LoRA uses significantly less trainable parameter and thus largely improves the computing and storage efficiency~\cite{hu2021lora}. 
As a variant method, Vector-based Random Matrix Adaptation (VeRA) freezes the two low-rank matrices via randomization and introduces two trainable scaling vectors, which again reduces the number of trainable parameters~\cite{kopiczko2023vera}.
As a result of randomized frozen matrix initialization, VeRA requires a higher rank for comparable performance, which increases the computation complexity.
AFLoRA points out that any reduction in VeRA's rank causes a drop in fine-tuning performance~\cite{liu2024aflora}.

Extending the state-of-the-art, 
we propose \project (\proj), a PEFT method that uses substantially few parameters than LoRA while showing close performance on various fine-tuning tasks.
Building upon VeRA, \proj also adopts a pair of randomized frozen projection matrices and a pair of scaling vectors $\vec{d}$ and $\vec{b}$ for adaptation.
We introduce an interpolation-based reparametrization mechanism: during the training process, \proj selectively reinitializes a part of the frozen matrices.
\proj achieves comparable or better performance than VeRA with a much lower rank.
With this approach, there is no need to perform gradient update on the projection matrices and 
the only trainable parameters are the pair of adaptation vectors.
\proj achieves state-of-the-art in terms of the number of trainable parameters and demonstrates efficient fine-tuning performance across various benchmarks.
Compared with LoRA, \proj achieves competitive fine-tuning performance while demanding 15x and 8x fewer parameters in GLUE and E2E benchmarks respectively.

Our main contributions are as follows:
\renewcommand{\thefootnote}{\arabic{footnote}}  
\begin{itemize}[leftmargin=*]
    \item We propose a novel parameter-efficient fine-tuning\footnote[1]{Code available at: \url{https://github.com/zhaojinm/UORA}} method with no inference latency, \project. Extending the state-of-the-art PEFT methods, \proj achieves performant results with much less trainable parameters.
    \item We compare our approach with LoRA, VeRA and other PEFT methods on natural language understanding (GLUE), and natural language generation (E2E) benchmarks, and compare instruction-tuning and image classification tasks.
    \item We conduct ablation studies to gain insights on various components of our approach and their impact on performance and efficiency.
\end{itemize}

\section{Related Work}

\subsection{Low-Rank Adaptation}

The state-of-the-art LoRA explores the intrinsic low dimension in language models, which involves finding a pair of low-rank decomposition matrices that capture the essential information of the weight updates during the training process~\cite{hu2021lora}. 
The decomposition contains a pair of projection-down and projection-down matrices, 
which significantly reduced the number of trainable parameters, and thus it largely increases the training and storage efficiency.
Additionally, these delta weights could be merged into the original model weights leading to zero inference latency.

A collection of methods extends the LoRA concept.
Dynamic LoRA (DyLoRA) allows for a range of training ranks, rather than a fixed one, to avoid the exhaustive search for the optimal rank~\cite{valipour2022dylora}.
Adaptive LoRA (AdaLoRA) adjusts the rank adaptively across different layers within a given budget~\cite{zhang2023adalora}.

Vector-Based Random Matrix Adaptation (VeRA) freezes the pair of projection matrices and introduces trainable scaling vectors for adaptation~\cite{kopiczko2023vera}.
AFLoRA reduces the decomposition rank by adaptively freezing the projection matrices at the beginning and eventually uses scaling vectors as VeRA~\cite{liu2024aflora}.
Our approach falls under the LoRA category and improves upon VeRA. 

Some research streams explore matrices decomposition into smaller chunks~\citep{shen-etal-2024-multimodal,ren-etal-2024-melora}, while others~\cite{shi-etal-2024-reslora} add residue modules.~\citet{dou-etal-2024-loramoe,zhang-etal-2024-milora} embed Mixture of Experts(MoE) architecture into LoRA. There are also various LoRA-based works~\citep{tian2024hydralora,ding-etal-2023-sparse,kim-etal-2024-ra}. Quantization techniques have also been added to further decrease memory usage~\cite{dettmers2023qlora,xu2024qalora,guo2024lqlora}.

\subsection{Parameter-Efficient Fine-Tuning}
PEFT refers to a series of methodologies that aim to achieve comparable fine-tuning performance at reduced costs in terms of time and storage requirements.
Compared to full fine-tuning, PEFT demonstrates advantages in low to medium resource setups and could be outperformed by full fine-tuning in high-resource setup~\cite{naveed2023comprehensive, chen2022revisiting}. In other words, depending on the location of the adaptation layers, PEFT methods are roughly divided into two categories, non-weight-based and weight-based.

Adapter tuning injects one or more trainable layers into the transformer blocks sequentially or in parallel~\cite{he2021towards, houlsby2019parameter, wang2022adamix,DBLP:conf/iccvw/FangYWZBFHLS23}; 
an adapter layer usually consists of downscaling, non-linearity and upscaling. The bottleneck with adapter tuning is that it introduces latency during inference and may reduce the parallelism in GPU computing operations.

Prompt tuning concatenates trainable prompt parameters withel embeddings for downstream tasks, and its derivatives address limitations in training instability and forgetting~\cite{liu2024gpt, lester2021power, liu2021p, jin2024disentanglingmemoryreasoningability}. 

Prefix tuning places a set of trainable vectors in the frozen transformer layers~\cite{li2021prefix, zhang2023towards}.
Prefix tuning shows non-linearity in performance against the number of trainable parameters.

Bias tuning, specifically BitFit, focuses on training the bias, a subset of the total trainable parameters in the model; however, it may appear limited when the training data are large~\cite{zaken2021bitfit}. 

There are some other efficient techniques such as data/token selection~\cite{dong2024towards,hu2024longreciperecipeefficientlong,sun2025uncertainty}, knowledge editing~\cite{deng2025everything,feng2025geoedit}, model pruning/compression~\cite{yu2024functionalfaithfulnesswildcircuit,zhou-etal-2023-modular}, distillation~\cite{feng2024continual,jia2024adversarial,wang-etal-2023-efficientvlm}, Linear Representation-Steering~\cite{bi2025llavasteeringvisualinstruction},chunk-wise gradient computation~\cite{li2025traininglongcontextllmsefficiently} and fine-tune partial layers of the model~\cite{fan-etal-2025-slam}.
\subsection{Parameter Modeling}
It has been widely studied to model the parameter distribution using randomized matrices for efficiency.
Given the model sparsity, randomly initialized weight matrices in neural networks appears to contain high performance sub-networks that requires less or even no training~\cite{frankle2018lottery, ramanujan2020s}.
It is empirically demonstrated that language models indeed possess low intrinsic dimension and it is efficient to employ randomized projection matrices~\cite{aghajanyan2020intrinsic}.
Meanwhile, there is a thread of research studying how information is distributed within LMs, and some works suggest that different neurons store different types of information~\cite{niu-etal-2022-bert,niu2024what}.

\section{Methodology}
\begin{table*}[h]
\centering
\begin{tabular}{l|r|r|r|r|r}
\toprule
& & \multicolumn{2}{c|}{\textbf{LoRA}}  & \multicolumn{2}{c}{\textbf{VeRA / \proj}} \\ 
Models&Rank   & \# Trainable & Required & \# Trainable & Required  \\ 
&   & Parameters & Bytes & Parameters & Bytes \\
\midrule
RoB\textsubscript{base} 
&1          & 36.8K   & 144KB   & 18.4K   & 72KB   \\ 
&16       & 589.8K  & 2MB     & 18.8K   & 74KB      \\ 
&256      & 9437.1K & 36MB    & 24.5K   & 96KB      \\ 
\midrule
RoB\textsubscript{large}    
&1        & 98.3K   & 384KB     & 49.2K     & 192KB   \\ 
&16       & 1572.8K & 6MB       & 49.5K     & 195KB    \\ 
&256      & 25165.8K& 96MB      & 61.4K     & 240KB    \\ 
\midrule
GPT-3
&1        & 4.7M    & 18MB    & 2.4M    & 9.1MB       \\ 
&16       & 75.5M   & 288MB   & 2.8M    & 10.5MB      \\ 
&256      & 1207.9M & 4.6GB   & 7M      & 33MB        \\ 
\bottomrule
\end{tabular}\caption{Theoretical memory required to store trained LoRA, VeRA, and \proj weights for RoBERTa Base (RoB\textsubscript{base}), and Large (RoB\textsubscript{large}), and GPT-3 models. We assume that all methods are applied on query and key layers of individual attention modules. Although \proj shares the same number of parameter for a given rank, \proj requires a lower rank by orders of magnitudes than VeRA in almost all scenarios.}
\label{tab:params}
\end{table*}

In this section, we present \project (\proj), a novel parameter-efficient fine-tuning approach inspired by state-of-the-art PEFT methods.

\subsection{Method Formulation}
LoRA exploits the low-intrinsic dimensionality in weight matrices and reparameterizes the weight update in fine-tuning with a pair of low-rank projection matrices.
Formally, for a pretrained weight matrix $W_0 \in \mathbb{R}^{d \times k}$, the weight updates after fine-tuning are decomposed by $\Delta W=BA$, where $B\in \mathbb{R}^{d \times r}, A\in \mathbb{R}^{r \times k}, r \ll min(d,k)$.
The forward pass is then modeled as:
\begin{equation}
h = W_0x + \Delta Wx = W_0x + BAx,
\end{equation}

To improve parameter efficiency, VeRA freezes the projection matrices $A$ and $B$ using randomized initialization,
and introduces trainable, scaling vectors $\vec{d}$ and $\vec{b}$:


\begin{equation}\label{eq:vera}
h = W_0x + \Delta Wx = W_0x + \Lambda_b B \Lambda_d A x, 
\end{equation}
$\Lambda_d$ and $\Lambda_b$ are the scaling vectors; they effectively scale or disable rows and columns in frozen matrices $A$ and $B$. 
Although the number of trainable parameters decreases, a higher rank is usually required for on-par fine-tuning performance. 

\proj adopts the weight update formula from Equation~\ref{eq:vera} in VeRA. However, it introduces a weight reinitialization mechanism, which has been shown to be highly efficient in neural network training \cite{pmlr-v187-zaidi23a}. After each iteration, we examine each entry in the scaling vector. If its magnitude falls below a threshold $\tau$ for a consecutive count of $k$ times, the corresponding column in matrix $A$ and the corresponding row in matrix $B$ are updated. To smooth the training process and avoid inconsistent performance drops, linear interpolation (LERP) with a factor $\alpha$ is applied between the old and new values. The interpolation-based reinitialization is achieved through the following formula. :

\begin{align} 
v_{rand} & = Random(v_{old})\\
v_{new} &= \alpha v_{old} + (1-\alpha) v_{rand} 
\end{align}


\subsection{Details}
Figure~\ref{fig.overview} right panel illustrates the pipeline of UoRA.

\paragraph{Orthogonal Uniform Initialization.}
The impact of initialization methods is studied in VeRA. The initialization method preserves matrix expressivity while maintaining a well-conditioned weight space for parameter modeling. However, in our work, we choose to use orthogonal initialization, based on its ability to improve gradient flow, ensure stability in deep networks, and accelerate convergence by preserving variance across layers~\cite{Hu2020Provable,ijcai2021p0355}. 

\paragraph{Dimension Pruning.}
Although the number of trainable parameters drastically reduces, VeRA heavily replies on the randomly initialized frozen matrices for reparameterization; thus it must adopt higher ranks for competitive performance in most cases. 
Rank 1024 in VeRA is commonly seen as oppose to rank 32 in LoRA.
Higher ranks punish not only the computation efficiency due to larger matrix multiplications but also storage efficiency.

The magnitude of scaling vectors indicates how important of the corresponding column or row in the frozen matrix A and B~\cite{sun2024a}.
Given a dimension in $\vec{d}$ whose magnitude is insignificant ($|\vec{d}|$ below threshold $\tau$ towards 0), then this dimension contributes less or even negligible to the weight update $\Delta W$.
Thus, a lower rank could be used for efficient reparameterization.

\paragraph{Interpolation-based Reinitialization.}


To enhance stability and prevent newly generated weights from becoming excessively large, we applied linear interpolation. This interpolation smooths the frozen weight updates, ensuring that the new weights do not deviate too far from the old ones; thus avoiding extreme values that could negatively impact training progress and consequently model performance. Linear interpolation is also widely used in machine learning~\cite{NEURIPS2024_6c49d2ad} and can be used to control the magnitude of parameter updates and prevent instability during training~\cite{iyer2024linear,berrada2020training}.

\subsection{Parameters}
Following the naming convention in the prior work, 
the number of trainable parameters in \proj is then 
\begin{equation}
    L_{tuned}\times(d_{model} + r).
\end{equation}
where $r$ denotes the rank, $d_{model}$ denotes the dimension of a given layer and $L_{tuned}$ denotes the number of fine-tuned layers.
VeRA share the same equation despite that its rank $r$ is typically larger the \proj's in most cases.
The number of trainable parameters in LoRA is computed as
\begin{equation}
    2\times L_{tuned} \times d_{model} \times r
\end{equation}
Benefied from the interpolation-based reinitialization mechanism, \proj can adopt a rank $r$ close to LoRA and much lower than VeRA;
Table~\ref{tab:params} compares the parameter efficiency in RoBERTa and GPT-3 models.


\section{Experiments}
In this section, we perform evaluations of \proj in the diverse domains, covering natural language understanding (NLU), natural language generation (NLG), instruction-tuning, and computer vision (CV). At last, we present an ablation study that sheds light on the effect of each component in our method.

\paragraph{Baselines.}
We compare \proj method with the state-of-the-art parameter-efficient fine-tuning methods. For fair, comprehensive comparison, we extend the experiment settings from prior works as much as we could. Baselines include:
\begin{itemize}[leftmargin=*]
    \setlength\itemsep{0em}
    \item Full fine-tuning (FFT) - All pretrained parameters in the model subjects to gradient updates for a downstream task; it offers the maximum flexibility but requires considerable resources.
    
    \item Adapter tuning - Following the naming convintion, Adapter\textsuperscript{H} injects adapters, with two fully connected layers and activation functions, between the attention module, multi-layer perceptron (MLP) module and the following residual connection~\cite{houlsby2019parameter}. 
    Adapter\textsuperscript{L} reduces the number of parameters by injecting only after MLP module and layer normalization operation~\cite{lin2020exploring}. 
    Similarly, Adapter\textsuperscript{P} (Adapterfusion) injects the adapter layer after the feed-forward network (FFN) module~\cite{pfeiffer2020adapterfusion}.
    Adapter\textsuperscript{D}, also known as AdapterDrop, selectively drops adapter layers for parameter efficiency~\cite{ruckle2020adapterdrop}. 
    
    \item Bitfit -  Bias tuning focuses on training bias vectors while freezing the rest of pretrained parameters in the model. 
    It may appear less efficient when training data is large~\cite{zaken2021bitfit}.
    
    \item LoRA - The state-of-the-art PEFT method that reparameterize the weight updates in form of $\Delta W =  BA$, where $B \in \mathbb{R}^{d\times r}, A\in \mathbb{R}^{r\times k}$ and rank $r$ is much smaller than model dimension for better paramter efficiency~\cite{hu2021lora}.
    
    \item AdaLoRA - Adaptive Low-Rank Adaptation lift the need to search for the rank $r$ via budget allocation strategy by employing singular value decomposition (SVD) adaptation~\cite{zhang2023adalora}.
    
    \item VeRA - Vector-based Random Matrix Adaptation further reduces the number of trainable parameters by freezing the projection matrices. It introduces a pair of trainable vectors, $b$ and $d$ for adaptation~\cite{kopiczko2023vera}. 
    \item Red -  Representation EDiting directly operates on representations at some layers via scaling and biasing operations~\cite{wu2024advancing}. 
    
\end{itemize}

\subsection{Natural Language Understanding}
\begin{table*}[h!]
\centering

\resizebox{\textwidth}{!}{
    \begin{tabular}{l|r|ccccccc}
    \toprule
\textbf{Model \& Method} & \textbf{\makecell{\# Trainable \\ Parameters}} & \textbf{\makecell{SST-2 \\ (Acc.)}} & \textbf{\makecell{MRPC \\(Acc.)}} & \textbf{\makecell{CoLA \\(MCC)}} & \textbf{\makecell{QNLI \\(Acc.)}} & \textbf{\makecell{RTE \\(Acc.)}} & \textbf{\makecell{STS-B \\(PCC)}} & \textbf{Avg.} \\
    \midrule
RoB\textsubscript{base}(FF)        & 125M  & 94.8   & 90.2   & 63.6   & 92.8   & 78.7   & 91.2   & 85.2 \\
RoB\textsubscript{base}(BitFit)    & 0.1M  & 93.7   & 92.7   & 62.0   & 91.8   & 81.5   & 90.8   & 85.4 \\
RoB\textsubscript{base}(Adpt\textsuperscript{D}) & 0.3M  & 94.2$_{\pm0.1}$ & 88.5$_{\pm1.1}$ & 60.8$_{\pm0.4}$ & 93.1$_{\pm0.1}$ & 71.5$_{\pm2.7}$ & 89.7$_{\pm0.3}$ & 83.0 \\
RoB\textsubscript{base}(Adpt\textsuperscript{D}) & 0.9M  & 94.7$_{\pm0.3}$ & 88.4$_{\pm0.7}$ & 62.6$_{\pm0.6}$ & 93.0$_{\pm0.2}$ & 75.9$_{\pm2.2}$ & 90.3$_{\pm0.4}$ & 84.2 \\
RoB\textsubscript{base}(LoRA)      & 0.3M  & 95.1$_{\pm0.2}$ & 89.7$_{\pm0.7}$ & 63.4$_{\pm1.2}$ & 93.3$_{\pm0.3}$ & 86.6$_{\pm0.7}$ & 91.5$_{\pm0.2}$ & 85.2 \\
RoB\textsubscript{base}(AdaLoRA)   & 0.3M  & 94.5$_{\pm0.2}$ & 88.7$_{\pm0.6}$ & 62.0$_{\pm0.4}$ & 93.1$_{\pm0.2}$ & 81.0$_{\pm0.6}$ & 90.5$_{\pm0.2}$ & 85.0 \\
RoB\textsubscript{base}(VeRA)    & 0.043M  & 94.6$_{\pm0.1}$ & 89.5$_{\pm0.5}$ & 65.6$_{\pm0.8}$ & 91.8$_{\pm0.2}$ & 78.7$_{\pm0.7}$ & 90.7$_{\pm0.2}$ & 85.2 \\
RoB\textsubscript{base}(RED) & 0.02M & 93.9$_{\pm0.3}$ & 89.2$_{\pm1.0}$ & 61.0$_{\pm3.0}$ & 90.7$_{\pm0.4}$ & 78.0$_{\pm2.1}$ & 90.4$_{\pm0.3}$ & 83.9\\
\rowcolor{blue!10} RoB\textsubscript{base}(\proj) & 0.019M & 94.2$_{\pm0.5}$ & 90.4$_{\pm0.5}$ & 65.4$_{\pm0.5}$  & 91.2$_{\pm0.2}$ & 87.1$_{\pm0.8}$ & 90.6$_{\pm0.2}$ & \textbf{86.5}\\
    \midrule
RoB\textsubscript{large}(FF)       & 356M  & 96.4   & 90.9   & 68.0   & 94.7   & 86.6   & 92.4   & 88.2 \\
RoB\textsubscript{large}(Adpt\textsuperscript{P}) & 3M    & 96.1$_{\pm0.3}$ & 90.2$_{\pm0.7}$ & 68.3$_{\pm1.0}$ & 94.8$_{\pm0.2}$ & 83.8$_{\pm2.9}$ & 91.9$_{\pm0.4}$ & 87.6 \\
RoB\textsubscript{large}(Adpt\textsuperscript{P}) & 0.8M  & 96.6$_{\pm0.2}$ & 89.7$_{\pm1.2}$ & 67.8$_{\pm2.5}$ & 94.8$_{\pm0.3}$ & 80.1$_{\pm2.9}$ & 91.9$_{\pm0.4}$ & 86.8 \\
RoB\textsubscript{large}(Adpt\textsuperscript{H}) & 6M    & 96.2$_{\pm0.3}$ & 88.7$_{\pm2.9}$ & 66.5$_{\pm4.4}$ & 94.7$_{\pm0.2}$ & 83.4$_{\pm1.1}$ & 91.9$_{\pm0.7}$ & 86.9 \\
RoB\textsubscript{large}(Adpt\textsuperscript{H}) & 0.8M  & 96.3$_{\pm0.2}$ & 87.7$_{\pm1.7}$ & 66.3$_{\pm2.7}$ & 94.8$_{\pm0.3}$ & 72.9$_{\pm2.9}$ & 91.7$_{\pm0.4}$ & 84.9 \\
RoB\textsubscript{large}(LoRA)     & 0.8M  & 96.2$_{\pm0.5}$ & 90.0$_{\pm1.0}$ & 68.2$_{\pm1.9}$ & 94.8$_{\pm0.3}$ & 85.2$_{\pm1.2}$ & 92.3$_{\pm0.5}$ & 87.8 \\
RoB\textsubscript{large}(VeRA)     & 0.061M  & 96.1$_{\pm0.1}$ & 90.9$_{\pm0.7}$ & 68.0$_{\pm0.8}$ & 94.4$_{\pm0.2}$ & 85.9$_{\pm0.7}$ & 91.7$_{\pm0.8}$ & 87.8 \\
\rowcolor{blue!10} RoB\textsubscript{large}(\proj) & 0.049M & 96.1$_{\pm0.2}$ & 92.2$_{\pm0.5}$  &  69.3$_{\pm0.5}$& 94.5$_{\pm0.5}$ & 87.0$_{\pm1.5}$ & 91.7$_{\pm0.5}$&\textbf{88.5}\\
    \bottomrule
    \end{tabular}

}    \caption{GLUE benchmark performance on RoBERTa Base (RoB\textsubscript{base}) and RoBERTa Large (RoB\textsubscript{large}). We report Matthew's correction coefficient (MCC) for CoLA, Pearson correlation coefficient (PCC) for STS-B, and accuracy (Acc.) for the rest tasks. Results for prior works are taken from~\citet{pmlr-v235-gao24o},~\citet{kopiczko2023vera} and~\citet{wu2024advancing}.}
\label{tab:glue}
\end{table*}

We evaluate \proj method on the General Language Understanding Evaluation (GLUE)~\cite{liu2019roberta} benchmark, a collection of natural language understanding (NLU) tasks. Typical tasks covers grammar, semantics, inference, and paraphrasing~\cite{wang2018glue}. 

\paragraph{Experiment Details.}
We employ RoBERTa Base (RoB\textsubscript{base}) and RoBERTa Large (RoB\textsubscript{large}) for the GLUE benchmark. 
With the aforementioned interpolation-based reinitialization methodology, \proj now can employ a much smaller rank $r$ than VeRA. 
For instance, \proj adopts a rank of 16 for MRPC task while VeRA requires a rank of 1024.
Detailed rank for individual tasks can be found in Appendix~\ref{sec:appendix}. 
We use orthogonal uniform initialization for both projection matrices $A$ and $B$ and we initialize the vector $\vec{d}$ and $\vec{b}$ with 0.1 and 0 respectively. 

Following prior work, \proj is applied to query and value projection matrices in the transformer blocks.
Similar to VeRA, we use separate learning rate for the head and \proj layers to mitigate the additional scaling hyperparameter introduced in LoRA.
Refer to Table~\ref{tab:param:glue} in Appendix~\ref{sec:appendix} for detailed hyperparameter settings.

We omit MNLI and QQP tasks in GLUE benchmark due to budget limitations.
We report the average of 5 runs with the best epoch evaluation result using randoms seeds in Appendix~\ref{sec:appendix}.

\paragraph{Results.}

Table~\ref{tab:glue} shows the results of \proj and baselines on GLUE benchmark.
\proj demonstrates competitive performance compared to LoRA while using approximately 15x less parameters. 
With an interpolation-based reinitialization mechanism, \proj is able to improve the performance on most of tasks compared to VeRA. 

\subsection{E2E Benchmark}
The E2E NLG benchmark, released by ~\citet{novikova2017e2e}, is an English dataset verbalized a set of 2-9 key-value attribute pairs in the restaurant domain, with more than 51K combinations of dialogues.

\begin{table*}[h!]
    \centering

    \begin{tabular}{@{}c|c|r|ccccc@{}}
        \toprule
\textbf{Model} & \textbf{Method}         & \textbf{\makecell{\# Trainable \\ Parameters}} & \textbf{BLEU} & \textbf{NIST} & \textbf{METEOR} & \textbf{ROUGE-L} & \textbf{CIDEr} \\
\midrule
        &FF\textsuperscript{*}   & 354.92M   &    65.95          & 8.52          & 45.95             & 69.13     & 2.35\\
        &Adpt\textsuperscript{H*} & 0.9M     &      64.31        & 8.29          & 44.91             & 67.72     & 2.28 \\
GPT2    &Adpt\textsuperscript{P*} & 0.8M     &      64.41        & 8.30          & 44.74             & 67.53     & 2.29 \\
Medium  &LoRA                    & 0.4M      &  67.14   & 8.65 &  46.05   & 69.50     &  2.41\\
        &VeRA                & 0.098M        &   66.34           & 8.52          & 45.84             & 69.39     & 2.39\\
        & RED       & 0.050M                & 64.62 & 8.33 & 45.14 & 67.46 & 2.25 \\
\rowcolor{blue!10}        &\proj                    & 0.051M    & 66.67             & 8.62          & 45.37             & 68.82     & 2.35 \\ 
\midrule
        &FT\textsuperscript{*}   & 774.03M   & 65.56        & 8.50 & 45.40 & 68.38 & 2.27  \\
        &Adpt\textsuperscript{H*} & 1.8M     & 65.94        & 8.46 & 45.78 & 68.65 & 2.23 \\
GPT2    &Adpt\textsuperscript{P*} & 1.5M     & 65.53        & 8.41 & 45.65 & 68.46 & 2.33 \\
Large   &LoRA                    & 0.77M     & 68.07        & 8.74 & 46.28 & 69.92 & 2.43 \\
        &VeRA                    & 0.18M     & 66.97        & 8.59 & 46.07 & 69.37 & 2.41 \\ 
        &RED                    & 0.09M      & 65.22        & 8.40 & 45.59 & 68.14 & 2.34 \\
\rowcolor{blue!10}        &\proj                  & 0.14M          & 67.77        & 8.60 & 46.17 & 69.05 & 2.40\\
        \bottomrule
    \end{tabular}    \caption{Performance of our methods on E2E test set via GPT2-medium and GPT2-large. Results with $^*$ are taken from~\citet{wu2024advancing}, rest are replicated by ourselves.}

\label{tab:e2e}

\end{table*}
\paragraph{Experiment Details.}
For natural language generation task, we evaluate \proj and other PEFT methods following the setup described in LoRA~\citep{hu2021lora}.
We fine-tune GPT-2 Medium and Large models~\citep{radford2019language} on the E2E benchmark dataset.
We adopt a rank of 32 for LoRA and \proj, 1024 for VeRA; the rest baselines are extended from prior work. 
We apply hyparameter tuning and details are depicted in the Appendix~\ref{sec:appendix}.

\paragraph{Results.}

As we can see from Table~\ref{tab:e2e}, \proj shows competitive performance compared to LoRA and VeRA, whose number of trainable parameters are 8x and 2x more.
Using similar number of parameters, \proj shows dominating performance in all tasks compared to RED.

\subsection{Instruction Tuning}
Pretrained models possess latent capabilities, which fine-tuning can unlock for specific downstream tasks. Building upon fine-tuning, instruction tuning introduces an additional instruction component into the training data, enabling models to better follow task-specific directives.
As a result, the instruction-tuned models become more robust and versatile in addressing prompt questions even with better reasoning capability.
To demonstrate \proj's generalization across various models
and its effectiveness on more complex tasks, we conduct evaluation on \textbf{Arithmetic Reasoning} benchmarks.

The evaluation dataset covers a wide range of arithmetic reasoning problems in various formats:
1) the \textbf{AddSub} dataset contains simple math word problems on addition and subtraction, typically involving one-step arithmetic reasoning~\cite{hosseini-etal-2014-learning};
2) the \textbf{MultiArith} dataset is a collection arithmetic word problems that require multiple operation steps like addition, subtraction, multiplication, and division to reach the solution~\cite{roy-roth-2015-solving};
3) the \textbf{SVAMP} dataset consists of arithmetic word problems up to grade 4 level, which is obtained by modifying  existing datasets to reduce annotation artifacts~\cite{patel-etal-2021-nlp};
4) the \textbf{SingleEq} dataset covers problems that are solvable using a single linear equation, emphasizing the translation from natural language to algebraic expressions~\cite{koncel-kedziorski-etal-2015-parsing}.

\begin{table*}[]
\centering

\begin{tabular}{l|c|r|cccc}
\toprule
\textbf{Model}          & \textbf{Method} & \textbf{\makecell{\# Trainable \\ Parameters}} & \textbf{AddSub} & \textbf{MultiArith} & \textbf{SVAMP}  & \textbf{SingleEq} \\ \hline
  & LoRA   & 8.4M     & 0.8051 & 0.9383     & 0.4286 & 0.7715   \\
LLaMA 7B  & VeRA   & 0.59M    & 0.8069 & 0.9300     & 0.3944 & 0.7303   \\
\rowcolor{blue!10}  & \proj   & 0.50M    & 0.8095 & 0.9133     & 0.4210 & 0.7047   \\ \hline
  & LoRA   & 8.4M     & 0.8962 & 0.9633     & 0.7539 & 0.9665   \\
LLaMA 3-8B & VeRA   & 0.59M    & 0.8886 & 0.9383     & 0.7027 & 0.9409   \\
\rowcolor{blue!10}  & \proj   & 0.50M    & 0.8911 & 0.9750     & 0.7100 & 0.9688 
\\
\bottomrule
\end{tabular}

\caption{Instruction tuning performance on arithmetic reasoning tasks using LLaMA 7B and LLaMA 3-8B models.}
\label{tab:mtbench}

\end{table*}
\paragraph{Experiment Details.}
We select LLaMA 7B and LLaMA 3-8B as the representative pre-trained models for instruction tuning.
For reproducibility purposes, we extend the experiment setting from previous work~\cite{hu-etal-2023-llm};
models are fine-tuned on the high-quality Math10K dataset,
obtained by selecting examples from arithmetic reasoning datasets and adding step-by-step rationales.
The fine-tuned models are then evaluated on AddSub, MultiArith, SVAMP, and SingleEq dataset.


All methods, LoRA, VeRA and \proj, are applied to all attention layers (namely $W_Q, W_K, W_V, W_o$).
A rank of 8 for LoRA is used, a rank 1024 for VeRA and rank 32 for \proj.
Detailed hyperparameter settings are disclosed in Appendix~\ref{sec:appendix}.

\paragraph{Results.}

Table~\ref{tab:mtbench} shows similar pattern observed in previous experiments.
We find that \proj maintains competitive performance in arithmetic reasoning in both LLaMA 7B and LLaMA 3-8B, despite using a minimal number of parameters shown in the column \# Trainable Parameters.
The \proj's extreme parameter efficiency could facilitate resource-constrained scenarios.

LoRA, with the most number of parameters and maximum flexibility, shows strong performance and sets the baseline for the rest methods. Despite using orders of magnitude fewer parameters, \proj closely follows LoRA in performance (e.g., in AddSub, LoRA scores 89.62\% and \proj closely follows with 89.11\% on LLaMA3-8B).

We are inspired by VeRA, which serves as our most immediate baseline; VeRA's scaling vectors are solely used to enable entries in frozen matrices. 
\proj, in contrast, could potentially offer better performance, because the interpolation-based reinitialization enhances the expressiveness of the frozen matrices. This not only reduces the number of trainable parameters but also offers improved fine-tuning performance as shown in Table~\ref{tab:mtbench}.


\subsection{Image Classification}
PEFT methods are as efficient in fine-tuning for image classification tasks.
Vision Transformer (ViT)~\cite{dosovitskiy2021an} successfully extends the attention mechanism to address image classification tasks.
In this section, we evaluate our method in ViT on common datasets.

\begin{table*}[t]
\centering

\label{tab:vera_lora}
\begin{tabular}{l|c|r|c c c c}
\toprule
\textbf{Model}    &\textbf{Method} & \textbf{\makecell{\# Trainable \\ Parameters}} & \textbf{CIFAR100} & \textbf{Food101} & \textbf{Flowers102} & \textbf{RESISC45} \\
\midrule
    &Head & --       & 79.02 & 70.85 & 56.86 & 40.84 \\
    &Full & 85.8M    & 95.00 & 85.25 & 98.05 & 93.28 \\
\textbf{ViT-B}
    &LoRA & 294.9K   & 96.78 & 86.49 & 99.09 & 94.29 \\
    &VeRA & 24.6K    & 95.38 & 84.78 & 98.03 & 92.01 \\
\rowcolor{blue!10}    &\proj & 19.2K       & 96.08 & 83.84 & 99.02 & 92.43 \\
\midrule
    &Head & --       & 82.22 & 72.63 & 63.26 & 50.47 \\
    &Full & 303.3M   & 95.44 & 88.64 & 99.02 & 95.23 \\
\textbf{ViT-L}
    &LoRA & 786.4K   & 96.56 & 85.38 & 99.00 & 94.60 \\
    &VeRA & 61.4K    & 95.78 & 85.08 & 98.04 & 93.54 \\
\rowcolor{blue!10}    &\proj & 50.7K       & 95.72 & 84.07 & 98.04 & 93.75 \\
\bottomrule
\end{tabular}\caption{Vision transformer results on image classification datasets with ViT Base (Vit-B) and Large (ViT-L). We report accuracy (\%) after 10 epochs.}
\label{tab:vit}
\end{table*}
\paragraph{Experiment Details.}
For evaluation, we adopt both ViT base and large models, pretrained on ImageNet-21K,
on a wide range of datasets, including RESIS45~\cite{7891544}, CIFAR100~\cite{krizhevsky2009learning}, FOOD101~\cite{bossard2014food}, FLOWER102~\citep{Nilsback08}.

We include full fine-tuning and training classification head only as baselines, 
as well as LoRA and VeRA.
Specifically, we apply all methods on the query and value layers in ViT models.
We followed the convention to use rank 8 for LoRA, 256 for VeRA and 32 for \proj.
Each run has 10 epochs, and detailed hyperparameter settings are listed in Appendix~\ref{sec:appendix}.

\paragraph{Results.}
As seen in Table~\ref{tab:vit},
\proj is able to outperform VeRA and achieve comparable performance to LoRA with fewer parameters.
For example, on CIFAR100, \proj reaches 96.08\% accuracy with ViT-B and 95.72\% with ViT-L, outperforming VeRA in both cases and approaching LoRA's performance, despite the number of trainable parameters is only 19.2K and 50.7K respectively.
This demonstrates the efficiency and generalization of \proj as a PEFT method for vision tasks.


\subsection{Ablation Study}

\subsubsection{Scalability}
\begin{figure}[!ht]
\centering
\includegraphics[width=\linewidth]{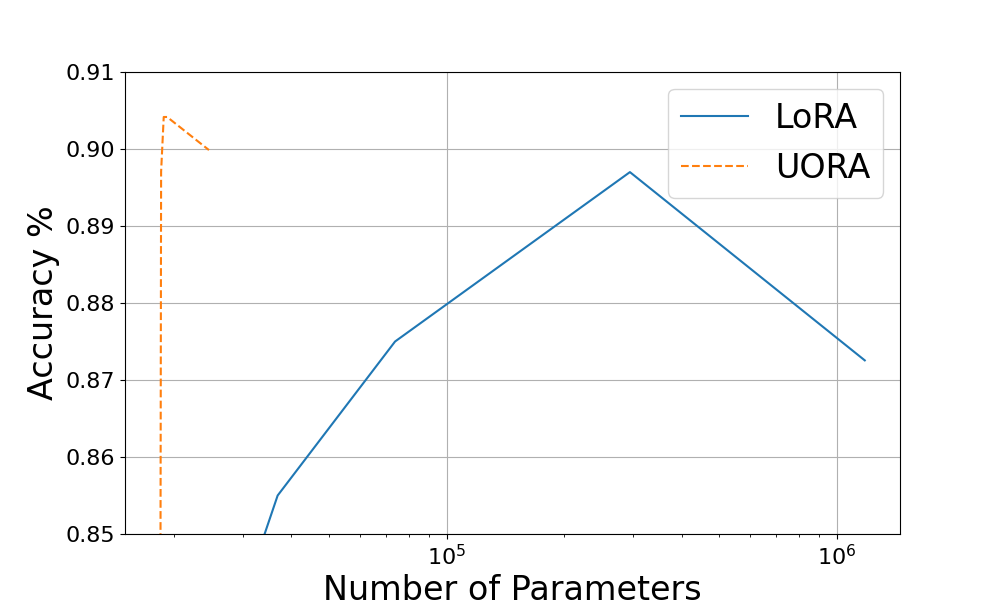}
\caption{Performance vs. number parameters of LoRA and \proj on MPRC in GLUE benchmark.}
\label{fig.scalability}
\end{figure}
We model the relation between fine-tune performance and the number of trainable parameters using RoBERTa Large model on MRPC in GLUE benchmark.
We adopt rank $r=\{1, 2, 4, 16, 64\}$ for LoRA and $r=\{1, 4, 16, 32, 256\}$ for \proj.
As shown in Figure~\ref{fig.scalability}, \proj demonstrates superior parameter efficiency, achieving comparable performance to LoRA while requiring fewer trainable parameters.
In higher ranks, both \proj and LoRA show performance slight degradation, which could be a sign for excessive amount of parameters.

\subsubsection{Impact of Initialization Method}
As aforementioned, we employ orthogonal uniform initialization for frozen projection matrices $A$ and $B$, as opposed to the Kaiming initialization in VeRA.
The randomized projection matrices are shared across all target layers.
The scaling vector $\vec{d}$ is initialized to 0.1 for efficient training as proved in VeRA, 
and and $\vec{b}$ is set to 0 such that the first forward pass is not affected, e.g., $\Delta W = \Lambda_b B \Lambda_d A x = 0$.

We conducted experiment to investigate the impact of initialization methods.
We implement different matrix initialization methods on top of VeRA (equivalent to \proj with reinitialization disabled) and test the performance for MRPC task in GLUE bench mark.
Table~\ref{tab:app:init} shows that both xavier and orthogonal uniform initialization shows slightly better performance than Kaiming uniform; random uniform intialization appears to be less effective in parameter modeling given its inefficacy in performance, which aligns with the observation in the VeRA paper.

\subsubsection{Impact of Threshold $\tau$, Count $k$, and Interpolation Factor $\alpha$}

The frequency of interpolation-based reinitialization depends on the threshold $\tau$ and the count $k$.
The extent of reinitialization depends on the interpolation factor $\alpha$.
After empirical study, we found that even though ($\tau, k, \alpha$) are correlated, the hyperparameter tuning can be simplified;
we depict practical tuning strategy in \ref{sec:appendix11}.
We conduct detailed ablation study in  
Appendix~\ref{sec:abstudy}.


\section{Conclusion}
In this paper, we introduce \proj, a novel PEFT method achieving both fine-tuning performance and parameter efficiency.
We extend VeRA and propose a simple yet effective interpolation-based reinitialization mechanism to improve the reparameterization efficiency.
Our evaluation validates that \proj is applicable to various domains including NLP and CV tasks.
Compared to the state-of-the-art LoRA, \proj demands 15x and 8x fewer parameters while demonstrating competitive performance in GLUE and E2E benchmarks respectively.

\section{Limitations}
We demonstrate the performance of our methods on various application domains.
There are many directions for future explorations.
While \proj updates the frozen matrices for improved performance,
more research could be done for more effective parameter modeling 
(e.g.,  guided by fine-grained heuristics).
Altough \proj demonstrates competitive performance,
there are remaining potentials to fuse with other PEFT methods
(e.g.,  alleviating the search of rank with singular value decomposition).  
Applying PEFT methods on extremely large language models, especially multimodal model, remains an open challenge.




\appendix

\section{Hyperparameters}
\label{sec:appendix}

We record hyperparameters below for reproducibility:
Table~\ref{tab:param:glue} for GLUE benchmark,
Table~\ref{tab:param:e2e} for E2E dataset,
Table~\ref{tab:param:it} for instruction turning, and
Table~\ref{tab:param:vit} for image classification using ViT.

\subsection{Tuning Strategy}
\label{sec:appendix11}

Hyperparameter tuning directly relates to the fine-tuning performance, and it can be tricky.
A good tuning strategy often saves users time during deployment.
The hyperparameter tuning process for UORA is relatively straightforward as follows.

\paragraph{Rank $r$}

We find rank 16 is most likely sufficient for smaller models like RoBERTa and GPT-2.
If computation resource allows, rank 32 often requires less tuning effort, and provides competitive performance out of box.

For large language models like llama and llama3, a minimum of rank 32 is needed.
We observe that a rank < 32 could lead to zero interpolation occurrences.
Higher ranks (e.g., $r$ > 256) could cause overparameterization and instability.

\paragraph{Learning rate $lr$}

UORA benefits from a learning rate similar to VeRA's, typically higher than LoRA's.
Given the small amount of trainable parameters, we find that 4e-2 is usually a good starting point.

\paragraph{Interpolation Factor $\alpha$}

Alpha = 0.7 provides a balanced trade-off.
A lower alpha < 0.5 introduces more noticeable turbulence in the training process
(as the learned scaling vectors takes longer to adapt to the largely varied frozen matrices).

\paragraph{Threshold $\tau$ and Count $k$}

Threshold $\tau$ and count $k$ together determine the number of reinitialization occurrences at a given index.
The goal is to control the number of reinitializations within an optimal range.
Too few reinitializations leads to similar performance as VeRA with a small rank;
too many reinitializations may disrupt the orthogonality of frozen matrices and training stability.

We start with $k=1$ and tune $\tau$ first.
We adjust the Tau value to avoid overly frequent reinitializations.
If tuning $\tau$ alone is insufficient to balance the number of reinitialized indices and total reinitializations, we then consider $k >= 2$, especially under noisier training regimes.

\begin{table*}[!htp]
\centering

\begin{tabular}{llcccccc}
\toprule
\textbf{Model} & \textbf{Hyperparameter} & \textbf{SST-2} & \textbf{MRPC} & \textbf{CoLA} & \textbf{QNLI} & \textbf{RTE} & \textbf{STS-B} \\
\midrule
               & Optimizer           & \multicolumn{6}{c}{AdamW} \\
               & Warmup Ratio        & \multicolumn{6}{c}{0.06} \\
               & LR Schedule         & \multicolumn{6}{c}{Linear} \\
               & Initialization & \multicolumn{6}{c}{Orthogonal uniform} \\
               & Initial Value of $d$ & \multicolumn{6}{c}{0.1} \\
               & Interpolation $\alpha$ & \multicolumn{6}{c}{0.7} \\
               & Count $k$          & \multicolumn{6}{c}{1} \\
               & Seed & \multicolumn{6}{c}{(0 42 88)} \\
\midrule
               & \# GPUs             & \multicolumn{6}{c}{1}       \\
               & Rank $r_q=r_v$          & 16    & 16    & 16    & 16    & 16    & 16    \\
               & Threshold $\tau$        & 1e-5    & 1e-5    & 1e-5    & 1e-5    & 8e-6   & 8e-6    \\
RoBERTa        & Epochs              & 60      & 60      & 80      & 25      & 160     & 40      \\
base           & Learning Rate (Head) & 4E-3    & 1E-2    & 1E-2    & 4E-3    & 1E-2    & 1E-2    \\
               & Learning Rate (\proj) & 4E-3    & 1E-2    & 1E-2    & 1E-2    & 4E-3    & 1E-2    \\
               & Max Seq. Len.       & \multicolumn{6}{c}{512}     \\
               & Batch Size          & \multicolumn{6}{c}{32}      \\
\midrule
               & \# GPUs             & \multicolumn{6}{c}{4}      \\
               & Rank $r_q=r_v$      & 32     & 32     & 32     & 32     & 32     & 32     \\
               & Threshold $\tau$          & 1e-5    & 1e-5    & 1e-5    & 8e-6    & 8e-6    & 8e-6    \\
RoBERTa        & Epochs              & 10      & 40      & 40      & 20      & 40      & 20      \\
large          & Learning Rate (Head) & 4E-3    & 1E-2    & 6E-3    & 2E-4    & 2E-3    & 5E-3    \\
               & Learning Rate (\proj) & 1E-2    & 1E-2    & 1E-2    & 1E-2    & 2E-2    & 1E-2    \\
               & Max Seq. Len.       & \multicolumn{6}{c}{128}     \\
               & Batch Size Per GPU  & \multicolumn{6}{c}{32}      \\
\bottomrule
\end{tabular}\caption{Hyperparameters for GLUE benchmark.}
\label{tab:param:glue}
\end{table*}

\begin{table*}[!htp]
\centering

\begin{tabular}{llcccc}
\toprule
\textbf{Model} & \textbf{Hyperparameter} & \textbf{LoRA} & \textbf{VeRA} & \textbf{RED} & \textbf{UORA} \\
\midrule
               & Optimizer              & \multicolumn{4}{c}{AdamW} \\
               & Warmup Steps           & \multicolumn{4}{c}{500} \\
               & Epochs                 & \multicolumn{4}{c}{5} \\
               & Label Smooth           & \multicolumn{4}{c}{0.0} \\
               & Batch Size             & \multicolumn{4}{c}{8}      \\
               & LR Schedule            & \multicolumn{4}{c}{Linear} \\
               & Seed                   & \multicolumn{4}{c}{(42 43 44)} \\
               & \# GPUs                & \multicolumn{4}{c}{1}       \\
               & Initial Value of $d$   &   -   & 0.1       & -         &0.1 \\
               & Initialization         &  -    & Kaiming       &   -   &Orthogonal \\
               &                        &       & uniform       &   -   &uniform \\
               & Threshold              & -    & -    & -         & 1                    \\
\midrule
GPT-2          & Rank  $r$             & 8    & 1024    & -    & 32        \\
Medium         & Learning Rate          & 6E-2    & 2E-2    & 6E-2    & 1E-2      \\
\midrule
GPT-2          & Rank $r$               & 8     & 1024     & -     & 32         \\
Large               & Learning Rate          & 4E-3    & 6E-3    & 2E-2    & 2E-2        \\
\bottomrule
\end{tabular}\caption{Hyperparameters for E2E benchmark.}
\label{tab:param:e2e}
\end{table*}

\begin{table*}[t!]

\centering
\begin{tabular}{l|ccc}
\toprule

        & LoRA & VeRA & \proj  \\
\midrule

\#GPU & \multicolumn{3}{c}{1}\\
Optimizer & \multicolumn{3}{c}{AdamW}\\
Warmup Ratio & \multicolumn{3}{c}{0.03}\\
Batch Size & \multicolumn{3}{c}{4}\\
Gradient Accumulation Steps & \multicolumn{3}{c}{4}\\
Epochs & \multicolumn{3}{c}{3}\\
LR Scheduler & \multicolumn{3}{c}{Cosine}\\
Weight Decay& \multicolumn{3}{c}{0.0}\\
Cutoff Length& \multicolumn{3}{c}{256}\\
Rank $r$ & 8 & 1024 & 32\\
Learning Rate & 3e-4 & 4e-2 & 4e-2\\
\bottomrule

\end{tabular}\caption{Hyperparameter setup for Instruction Tuning. LoRA alpha = 16. \proj $\tau $ = 5e-5; k = 1.}
\label{tab:param:it}
\end{table*}





\begin{table*}[t!]
\centering

\begin{tabular}{cl|cccc}
\toprule

\textbf{Model} & \textbf{Hyperparameter} & \textbf{CIFAR100} & \textbf{Food101} & \textbf{Flowers102} & \textbf{RESISC45}  \\
\midrule

& Weight Decay & \multicolumn{4}{c}{0.0}\\
& Optimizer           & \multicolumn{4}{c}{AdamW} \\
& LR Schedule           & \multicolumn{4}{c}{Linear} \\
& Rank $r$           & \multicolumn{4}{c}{32} \\
& epoch                   & \multicolumn{4}{c}{10} \\
& Seed  & \multicolumn{4}{c}{(42,43,44)} \\
\midrule

\multirow{5}{*}{Base} 
& LR-head (Head only)   & 5e-5  & 5e-5  & 5e-5  & 5e-5  \\
& LR (Full)             & 5e-5  & 5e-5  & 5e-5  & 5e-5  \\
& LR-head (LoRA)        & 4e-3  & 4e-3  & 4e-3  & 4e-3  \\
& LR (LoRA)             & 4e-3  & 4e-3  & 4e-3  & 4e-3  \\
& LR-head (VeRA)        & 4e-3  & 4e-3  & 4e-3  & 4e-3  \\
& LR (VeRA)             & 4e-2  & 4e-2  & 4e-2  & 4e-2  \\
& LR-head (\proj)       & 4e-3  & 4e-3  & 4e-3  & 4e-3  \\
& LR (\proj)            & 4e-2  & 4e-2  & 4e-2  & 4e-2  \\
\midrule

\multirow{5}{*}{Large} 
& LR-head (Head only)   & 5e-5  & 5e-5  & 5e-5  & 5e-5  \\
& LR (Full)             & 5e-5  & 5e-5  & 5e-5  & 5e-5  \\
& LR-head (LoRA)        & 4e-3  & 4e-3  & 4e-3  & 4e-3  \\
& LR (LoRA)             & 4e-3  & 4e-3  & 4e-3  & 4e-3  \\
& LR-head (VeRA)        & 4e-3  & 4e-3  & 4e-3  & 4e-3  \\
& LR (VeRA)             & 4e-2  & 4e-2  & 4e-2  & 4e-2  \\
& LR-head (\proj)       & 4e-3  & 4e-3  & 4e-3  & 4e-3  \\
& LR (\proj)            & 4e-2  & 4e-2  & 4e-2  & 4e-2  \\\bottomrule

\end{tabular}\caption{Hyperparameter setup for ViT on the image classification benchmarks.}
\label{tab:param:vit}
\end{table*}


\newpage
\section{Ablation Study}
\label{sec:abstudy}
This section depicts the initialization comparison and sensitivity analysis of \proj-specific hyperparameters ($\tau, k, \alpha$).


\begin{table}[h]
    \centering
    \begin{tabular}{r|r}
\toprule
Initialization      & MRPC \\ \midrule
Xavier Uniform      & 88.72 \\
Orthogonal Uniform  & 88.72 \\
Kaiming Uniform     & 87.99 \\
Random Uniform      & 72.05 \\
\bottomrule
    \end{tabular}
    \caption{Impact of Matrix Initialization}
    \label{tab:app:init}
\end{table}

\paragraph{Threshold $\tau$.}
$\tau$ controls the magnitude threshold to trigger reinitialization.
If a dimension in $\vec{d}$ becomes lower than $\tau$, 
it is considered as less efficient in modeling the weight distribution of target layers. 
We conducted a focused experiment on LLaMA 7B on the AddSub dataset to evaluate the impact of the threshold $\tau$.

As shown in Table~\ref{tab:app:threshold}, a higher value of $\tau$ leads to more frequent reinitialziation; 
excessive occurrences could introduce performance degradation.
$\tau$ = 1e-4 provides the best performance in this experiment.
As $\tau$ decreases further, the performance begins to drop, which highlights the importance of a balanced reinitialization frequency.


\begin{table}[h]
    \centering
    \begin{tabular}{c|c}
    \toprule
    Threshold $\tau$ & AddSub \\ \hline
    5e-4          & 0.6759 \\
    1e-4          & 0.8095 \\
    5e-5          & 0.7620 \\
    1e-5          & 0.7063 \\
    \bottomrule
    \end{tabular}
    \caption{Impact of Threshold $\tau$}
    \label{tab:app:threshold}
\end{table}

\paragraph{Count $k$.}
Count $k$ controls the number of training steps where a dimension is below $\tau$ before reinitialization triggers.
A higher $k$ reduces the frequency of reinitialization and makes \proj conservative about the current frozen matrices.
Special values like $k = 0$ disables \proj reinitialization mechanism; another case is that when $k$ is large enough that the consecutive count is never reached.
We conduct empirical study to make sense of the appropriate range for $k$.
We observe that given a tuned threshold $\tau$, $k \in \{1, 2, 3\}$ shows promising performance.

\begin{table}[h]
    \centering
    \begin{tabular}{c|c}
\toprule
Count $k$      & MRPC \\ \midrule
0       & 89.95 \\
1       & 90.44 \\
2       & 89.99 \\
3       & 89.95 \\
4       & 89.95 \\
\bottomrule
    \end{tabular}
    \caption{Impact of Count $k$}
    \label{tab:app:count}
\end{table}

\paragraph{Interpolation Factor $\alpha$.}
$\alpha$ controls the magnitude of change when modifying a frozen matrix during reinitialization.
A higher $\alpha$ leads to more stabilized training process at a cost of inefficient parameter modeling.
When $\alpha = 1$, \proj can be viewed a close variant of VeRA where the initialization method adopts orthogonal uniform.
Table~\ref{tab:app:alpha} shows the impact of interpolation factor $\alpha$ on the E2E dataset.
We observe that $\alpha=0.7$ provides the best trade-off between performance and stability as it preserves a better approximate orthogonality than smaller $\alpha$'s.

\begin{table}[h]
\centering
\small
\begin{tabular}{r|rrrrr}
\toprule
$\alpha$    & 0.3       & 0.5       & 0.7       & 1         \\ \midrule
BLEU        & 63.90     & 63.75     & 66.67     & 65.78     \\
NIST        & 8.35      & 8.25      & 8.61      & 8.49      \\
METEOR      & 44.96     & 45.20     & 45.37     & 45.58     \\
ROUGE\_L    & 67.64     & 68.51     & 68.82     & 69.01     \\
CIDEr       & 2.30      & 2.32      & 2.35      & 2.32      \\
\bottomrule
\end{tabular}
\caption{Sensitivity analysis of interpolation factor $\alpha$.}
\label{tab:app:alpha}
\end{table}

\newpage
\section{Instruction Tuning Examples}
\label{sec:it_example}
Table~\ref{tab:it:eg} shows an example of the multi-turn questions in MT-Bench.
Table~\ref{tab:it:answer} shows the LoRA answers and \proj for the same question.

\begin{table*}[]
    \centering
    \begin{tabular}{|p{0.1\textwidth}|p{0.8\textwidth}|}
        \toprule
        Q1  & Extract the following information from the presented texts: The name of the book, the author, the main character, the year of publication. Output in the format of “main character, book, author, year of publication”, one book per line. a) In the realm of wizarding literature, a true standout is the work of J.K. Rowling. One of her books that left an indelible mark is ‘Harry Potter and the Philosopher’s Stone’. This iconic tale, published in 1997, tells the story of Harry, a young orphan who discovers his magical abilities on his 11th birthday. Soon, he finds himself at the Hogwarts School of Witchcraft and Wizardry, a place teeming with magic and adventure, located somewhere in Scotland. b) The magic of Middle-earth has entranced readers worldwide, thanks to the brilliance of J.R.R. Tolkien. In one of his seminal works, ‘The Lord of the Rings: The Fellowship of the Ring’, published in 1954, we meet Frodo Baggins, a brave hobbit tasked with the perilous quest of destroying the One Ring. The epic journey takes him from the peaceful Shire to the tumultuous regions of Middle-earth. c) In a galaxy far, far away, the imagination of L.E. Starlighter gives us ‘The Prism Galaxy Chronicles: The Awakening of the Starcaster’. Published in 2028, the story is about Zylo, a humble spaceship mechanic, who unexpectedly discovers he’s a Starcaster - a rare individual with the power to manipulate stardust. Set against the backdrop of an interstellar empire in turmoil, Zylo’s destiny unfolds on numerous alien worlds, each with its unique cosmic charm.  \\ 
        \hline
        Q2  & Reformulate your earlier reply, output it in JSON format and only include books published after 1980. \\
        \bottomrule
    \end{tabular}
    \caption{No. 131 question in MT-Bench}
    \label{tab:it:eg}
\end{table*}

\begin{table*}[]
    \centering
    \begin{tabular}{|p{0.1\textwidth}|p{0.4\textwidth}|p{0.4\textwidth}|}
        \toprule
            & LoRA  & \proj \\ \hline
        Q1  & The name of the book is 'Harry Potter and the Philosopher's Stone', the author is J.K. Rowling, the main character is Harry Potter, and the year of publication is 1997. 
        & Here are the answers to your question:

a) The main character is Harry Potter, the book is 'Harry Potter and the Philosopher's Stone', the author is J.K. Rowling, and the year of publication is 1997.

b) The main character is Frodo Baggins, the book is 'The Lord of the Rings: The Fellowship of the Ring', the author is J.R.R. Tolkien, and the year of publication is 1954.

c) The main character is Zylo, the book is 'The Prism Galaxy Chronicles: The Awakening of the Starcaster', the author is L.E. Starlighter, and the year of publication is 2028.
        \\ \hline
        Q2  & 
\{
"books": [
    
    \{
    
        "name": "Harry Potter and the Philosopher's Stone",
        
        "author": "J.K. Rowling",
        
        "main\_character": "Harry Potter",
        
        "year\_of\_publication": 1997
        
    \},
    
    \{
        
        "name": "The Lord of the Rings: The Fellowship of the Ring",
        
        "author": "J.R.R. Tolkien",
        
        "main\_character": "Frodo Baggins",
        
        "year\_of\_publication": 1954
        
    \},
    
    \{
        
        "name": "The Prism Galaxy Chronicles: The Awakening of the Starcaster",
        
        "author": "L.E. Starlighter",
        
        "main\_character": "Zylo",
        
        "year\_of\_publication": 2028
        
    \}
    
]
\}
        & Here is the reformulated answer:

a) \{
"main character": "Harry Potter",
"book": "Harry Potter and the Philosopher's Stone",
"author": "J.K. Rowling",
"year of publication": "1997"
\}

b) \{
"main character": "Frodo Baggins",
"book": "The Lord of the Rings: The Fellowship of the Ring",
"author": "J.R.R. Tolkien",
"year of publication": "1954"
\}

c) \{
"main character": "Zylo",
"book": "The Prism Galaxy Chronicles: The Awakening of the Starcaster",
"author": "L.E. Starlighter",
"year of publication": "2028"
\}
        \\
        \bottomrule
    \end{tabular}
    \caption{Example answers to MT-Bench questions. Answers formatted according to newline character in the model output.}
    \label{tab:it:answer}
\end{table*}

\end{document}